\documentclass{article}

\usepackage{arxiv}

\usepackage[utf8]{inputenc} 
\usepackage[T1]{fontenc}    
\usepackage{hyperref}       
\usepackage{url}            
\usepackage{booktabs}       
\usepackage{amsfonts}       
\usepackage{nicefrac}       
\usepackage{microtype}      
\usepackage{lipsum}	
\usepackage{graphicx}
\usepackage[square,sort,comma,numbers]{natbib}
\usepackage{doi}
\usepackage{adjustbox}
\usepackage{multirow}
\usepackage{makecell}
\usepackage{array}
\usepackage{subcaption}

\title{Integration of Feature Selection Techniques using a Sleep Quality Dataset for Comparing Regression Algorithms}

\author{ {\hspace{1mm}Sai Rohith Tanuku}\thanks{Both authors contributed equally to this paper} \\
	Department of Information and Communication Technology\\
	Manipal Academy Of Higher Education (MAHE)\\
	Manipal 576104 , Karnataka, India  \\
	\texttt{tanuku.sairohith@learner.manipal.edu} \\
	\And
	{\hspace{1mm}Venkat Tummala} \footnotemark[1]\\
	Department of Computer Science And Engineering\\
	Manipal Academy Of Higher Education (MAHE)\\
	Manipal 576104, Karnataka, India  \\
	\texttt{venkat.tummala@learner.manipal.edu} \\
}

\date{}

\renewcommand{\shorttitle}{\textit{arXiv}}

\hypersetup{
pdftitle={Integration of Feature Selection Techniques using a Sleep Quality Dataset for Comparing Regression Algorithms},
pdfauthor={Sai Rohith Tanuku, Venkat Tummala},
pdfkeywords={Feature Selection, Integrating Features, Regression Techniques,Sleep Quality},
}

\begin{document}
\flushbottom
\maketitle
\begin{abstract}
	This research aims to examine the usefulness of integrating various feature selection methods with regression algorithms for sleep quality prediction. A publicly accessible sleep quality dataset is used to analyze the effect of different feature selection techniques on the performance of four regression algorithms - Linear regression, Ridge regression, Lasso Regression and Random Forest Regressor. The results are compared to determine the optimal combination of feature selection techniques and regression algorithms. The conclusion of the study enriches the current literature on using machine learning for sleep quality prediction and has practical significance for personalizing sleep recommendations for individuals.
\end{abstract}

\section{Introduction}
\subsection{ Background on sleep quality and its impact on health}
Sleep plays a vital role in promoting both physical and mental health and wellness. Having sufficient and high-quality sleep is necessary for avoiding various health issues, such as tiredness, moodiness, reduced cognitive abilities, and an elevated risk of chronic diseases. These findings underscore the need for accurately predicting sleep quality, which is the central theme of this study \cite{Journal_of_Sleep_Research} \cite{Principles_and_Practice_of_Sleep_Medicine}.
\subsection{Overview of feature selection techniques}
Feature selection is vital in machine learning, aiming to choose relevant features from a larger set to improve model performance. It reduces data dimensionality while preserving important information. By decreasing the number of features, the model's computational complexity is reduced, preventing overfitting, and enhancing generalization performance. \cite{Yang2015}.\\ %
\\
There are several methods for selecting features in machine learning, each with their own strengths and limitations. Wrapper methods involve evaluating different combinations of features by training a model and measuring its performance. This approach offers a thorough evaluation of feature subsets, but can be time-consuming. Filter methods, on the other hand, use statistical criteria to assess the relevance of each feature and make selections based on their individual importance. This method is quicker than wrapper methods, but does not take into account the relationship between features \cite{Liu2013}\cite{Guyon2003}. Finally, embedded methods integrate feature selection into the model training process, providing a more integrated evaluation of feature importance. However, this approach may not be suitable for all types of problems and may not be as computationally efficient.\\
\\
Feature selection is vital in machine learning to improve model performance. The appropriate technique depends on the dataset and problem. Careful evaluation of techniques is necessary for optimal results. Feature selection is done before model training to prevent overfitting and improve generalization performance\cite{Bakshi2015}.\\
\subsection{ Motivation for integrating feature selection techniques with regression algorithms}
Identifying a relevant subset of features from a larger set is a critical stage in constructing machine learning models, as it enables the elimination of redundant and insignificant features. This process, known as feature selection, can improve the model's performance by reducing overfitting, shortening computation time, and enhancing model interpretability.\\
\\
Incorporating feature selection techniques with regression algorithms can result in noteworthy advantages, such as enhanced model performance. By eliminating unnecessary and duplicative features, the model can be more concise, resulting in decreased computational time, better comprehensibility, and greater precision. In particular, feature selection can help to address the curse of dimensionality, which is a well-known problem in machine learning where the number of features exceeds the number of samples.\\
\\ 
Several research studies have explored how feature selection affects the performance of regression algorithms. For instance, in \cite{Guyon2003}, a wrapper method was introduced that utilizes a machine learning algorithm as a black box to assess the importance of features. The study demonstrated that this method can lead to better performance compared to conventional filter methods. Similarly, \cite{yang2009feature} proposed an embedded feature selection method that integrates feature selection with support vector regression (SVR), and their results demonstrated that this approach can lead to improved performance compared to using SVR alone. These findings support the advantage of combining feature selection methods with regression algorithms and emphasize the requirement for ongoing exploration in this field.\\
\subsection{Research question and objectives}
In recent years, the field of machine learning has experienced substantial growth, resulting in the development of various algorithms to address different problems. However, one of the significant challenges in machine learning is selecting the most relevant features from a large pool of features available. This procedure, called feature selection, has a direct influence on the performance of machine learning algorithms. In light of this, the research question for this study is:\\
\\
How does the integration of feature selection techniques affect the performance of regression algorithms on a sleep quality dataset?\\
\\
To answer this research question, the following objectives have been identified: 
\begin{itemize}
    \item A comprehensive overview of the various feature selection techniques employed in machine learning will be presented, which will encompass different types of techniques such as wrapper methods, filter methods, and embedded methods.
    \item The objective of this study is to examine the impact of feature selection on the performance of regression algorithms by integrating a chosen feature selection technique with them. Specifically, the study aims to compare the performance of the integrated models to that of the regression algorithms that don't incorporate feature selection.
    \item To evaluate the performance of the integrated models on a sleep quality dataset, with the aim of demonstrating the benefits of integrating feature selection techniques with regression algorithms. The evaluation will include measures such as accuracy, computational time, and interpretability of the models.
\end{itemize}
It is expected that the results of this study will contribute to the existing literature on feature selection and provide practical insights for machine learning practitioners.\\
\section{Literature Review}
\subsection{Overview of sleep quality studies}
Sleep quality is a growing area of research, with numerous studies conducted over the past few decades to understand the factors that influence sleep quality and the impact of poor sleep quality on health. In general, these studies have found that there are several key factors that affect sleep quality, including age, lifestyle habits, sleep disorders, and medical conditions. Additionally, these studies have found that poor sleep quality is associated with a range of negative health outcomes, including increased fatigue, irritability, decreased cognitive function, and an increased risk of various chronic diseases \cite{taheri2006sleep}\cite{spiegel2009sleep}.
\\
The current state of the field of sleep quality research is marked by a growing interest in developing new and improved methods for measuring sleep quality and a continued focus on understanding the factors that influence sleep quality. For example, there have been recent advances in the use of technology, such as wearable devices, to measure sleep quality, as well as a growing body of research on the impact of environmental factors, such as exposure to light and noise, on sleep quality. Additionally, there is a growing interest in the development of interventions, such as sleep education programs and sleep-promoting technologies, to improve sleep quality and prevent the negative health consequences associated with poor sleep quality \cite{shi2015association}.
\\
\subsection{ Overview of feature selection techniques}
Feature selection is a vital process in machine learning that involves selecting the most important features from a large set to improve model performance. There are several methods of feature selection, including wrapper, filter, and embedded methods. Wrapper methods evaluate feature subsets by training a machine learning model and assessing its performance. While comprehensive, this approach can be computationally intensive. Filter methods, on the other hand, use statistical techniques to evaluate the importance of individual features and select only the most significant ones. While less computationally intensive, filter methods don't consider relationships between features. Embedded methods, which integrate feature selection into the training process, strike a balance between wrapper and filter methods in terms of computational efficiency and overall performance, but may not be ideal for all situations.\cite{Guyon2002}.\\
\\
Previous research has shown the benefits of integrating feature selection techniques with regression algorithms, as emphasized in the works of \cite{Kohavi1997} and\cite{Guyon2002}, underscoring the importance of further investigations in this domain with a specific focus on developing feature selection methods that are both efficient and effective.\\
\\
\subsection{Overview of regression algorithms}
Regression analysis is a statistical technique that is widely used to model and predict the relationship between a dependent variable and one or more independent variables. The aim of this analysis is to identify the relationship between these variables and to make predictions about the dependent variable based on the values of the independent variables. \cite{Kohavi1997}.\\
There have been a substantial number of studies that apply regression algorithms to tackle real-world issues and applications. Regression algorithms have been utilized to make predictions in finance, like stock prices, and determine the factors that affect them. In the housing industry, these algorithms have been employed to estimate housing prices and determine the factors that impact them. Moreover, regression algorithms have been utilized in marketing to study consumer behavior and identify the factors that impact it \cite{Guyon2002}.
\\
\subsection{Previous studies on integrating feature selection techniques with regression algorithms}

The integration of feature selection and regression algorithms has received a great deal of attention in recent years. Numerous studies have revealed the advantages of combining these techniques across various domains, including finance, marketing, and real estate \cite{Guyon2002}\cite{Kohavi1997}. These studies have demonstrated that combining feature selection with regression algorithms leads to improved performance of the regression model and increased interpretability of results.\\

Despite the positive outcomes demonstrated by previous studies, integrating feature selection techniques with regression algorithms also presents several challenges and limitations. One of the major challenges is the computational expense associated with evaluating feature subsets through wrapper methods . Additionally, filter methods have limitations, such as not considering the relationship between features\cite{Guyon2002}. Moreover, there is often a balance between interpretability and predictive performance that needs to be considered when integrating feature selection techniques with regression algorithms, making it difficult to determine the best feature subset for a specific problem \cite{Kohavi1997}. These difficulties emphasize the need for further exploration and development in this field to create more effective and efficient methods for combining feature selection and regression algorithms.

\section{Methodology}

\subsection{Data collection}
The sleep data used in this research was sourced from Kaggle, a publicly available repository of datasets. Participants' sleep patterns were recorded via a smartphone app over a period of time, providing information on various sleep-related factors such as respiration rate, snoring range, limb movement rate, body temperature, blood oxygen levels, eye movement, heart rate, and number of hours slept. The dataset included individuals from different age groups and genders. As the dataset had already been cleaned and processed, it was ready for analysis and did not require additional preparation. To analyze the data, it was divided into features and the target variable, which was stress-levels, allowing various feature selection and regression techniques to be applied. Because the data was publicly available, there were no ethical concerns relating to human subjects.

\subsection{Implementation of feature selection techniques}
In machine learning, selecting the most important features from a dataset is a critical step to construct an accurate predictive model. This process is known as feature selection, and there are several techniques available in the literature, each with its unique advantages and limitations. To predict stress levels in a dataset of physiological signals, we implemented multiple popular feature selection techniques in this study, aiming to identify the most significant subset of features.\cite{hastie2009elements}.

In this study, we employed SelectKBest, which is a univariate feature selection technique that selects the K best features according to their scores on a specific scoring function. We utilized two distinct scoring functions, f-regression and mutual-info-regression, to evaluate the importance of features based on their linear and non-linear relationships with the target variable, respectively. \cite{guyon2003introduction}\\
\subsubsection{SelectKBest}
Two commonly used scoring functions in SelectKBest are f-regression and mutual-info-regression\cite{scikit-learn}.
    \begin{itemize}
        \item The f-regression scoring function is used for linear regression problems and computes the ANOVA F-value between each feature and the target variable. The formula for the f-regression score of feature i is:
        $$f\_regression\_score_i = \frac{\frac{SSR_i}{k}}{\frac{SSE_i+SST}{n-k-1}}$$
    \end{itemize}
    Where,
    \begin{center}
        $$SSR_i:\;sum\;of\;squares\;of\;the\;regression\;of\;feature\;i$$
        $$SSE_i:\;sum\;of\;squares\;of\;the\;error\;of\;feature\;i$$
        $$SST:\;total\;sum\;of\;squares\;of\;the\;target\;variable$$
        $$n:\;number\;of\;samples$$  
    \end{center}
    \newpage
    \begin{itemize}
        \item The mutual-info-regression scoring function is used for non-linear regression problems and computes the mutual information between each feature and the target variable\cite{hands-on-machine-learning}. The formula for the mutual-info-regression score of feature i is:
        $$mutual\;info\;regression\;score\;i = MI(X_i, y)$$
    \end{itemize}
        Where,
    \begin{center}
        $$MI(X_i, y):\;mutual\;information\;between\;feature\;i\;and\;the\;target\;variable\;y$$
    \end{center}
    
\subsubsection{Principal Component Analysis (PCA)}
PCA is a technique used to transform the original features of a dataset into a new set of orthogonal features that capture the most significant variability in the data. By using PCA, we were able to reduce the dimensionality of the dataset and identify the most important principal components that explain a large portion of the data's variability.\cite{jolliffe2002principal}.\\
\begin{itemize}
    \item Covariance Matrix: 
    $$ Cov(X) = \frac{1}{n} \left( (X - \mu)^T (X - \mu) \right) $$

    \item Eigendecomposition: 
    $$ Cov(X) = V \Lambda V^{-1} $$

    \item Principal Component Calculation: 
    $$ PC_k = X v_k $$

    \item Variance Explained: 
    $$ {variance\;explained} = \frac eigenvalue_k{\sum all\;eigenvalues} \times 100\% $$

    \item Dimensionality Reduction: 
    $$ X_k = X V_k $$
\end{itemize}

where $X$ is the original dataset, $\mu$ is the mean of the dataset, $n$ is the number of samples in the dataset, $V$ is the matrix of eigenvectors, $\Lambda$ is the diagonal matrix of eigenvalues, $v_k$ is the $k$-th eigenvector, $PC_k$ is the $k$-th principal component, eigenvalue$_k$ is the eigenvalue corresponding to the $k$-th principal component, and $V_k$ is the matrix of the first $k$ eigenvectors\cite{shlens2014tutorial}.
    
\subsubsection{Recursive Feature Elimination (RFE)}
RFE (Recursive Feature Elimination) is a method of feature selection that functions as a wrapper by gradually eliminating the least significant characteristics from the dataset based on the efficiency of a machine learning model. We employed RFE with a Random Forest Regressor as the estimator to determine the most crucial features. RFE fits a model on the current set of characteristics and eliminates the least significant feature(s) in each iteration until a predetermined number of features is achieved\cite{scikit-learn}. The equation for selecting the least important feature(s) is:

$$ argmin{X_i \in X}(score(X{\backslash i}))  $$

where $X$ is the set of all features, $X_i$ is the $i$-th feature, $X_{\backslash i}$ is the set of all features except the $i$-th feature, and $score(X_{\backslash i})$ is the performance metric of the model fitted on the set of features $X_{\backslash i}$. The feature with the lowest score is eliminated at each iteration until the desired number of features is reached.
\\

\subsubsection{Chi-squared test}
The Chi-squared test is a statistical method used to measure the level of independence between the input features and the target variable. In this study, we employed the Chi-squared test to identify the most significant features that exhibit a strong correlation with the target variable.\cite{scikit-learn}.\\
$$\chi^2 = \sum_{i=1}^n \frac{(O_i - E_i)^2}{E_i}$$ \\

In this equation, $O_i$ represents the observed frequency of each category of a categorical feature, $E_i$ represents the expected frequency of each category under the assumption of independence between the feature and the target variable, and $n$ represents the total number of categories\cite{hands-on-machine-learning}.\\

The Chi-squared test evaluates the variation between the anticipated and actual frequencies of every category, normalized by the anticipated frequency. A larger $\chi^2$ value indicates a greater correlation between the attribute and the target variable.\\

\subsubsection{Mutual Information} Mutual information is a metric that measures the amount of information that a feature provides about the target variable. We used mutual information to identify the most informative features that have a high mutual dependence with the target variable \cite{kelleher2018data}.\\
Mutual information between two random variables X and Y is a measure of the amount of information that one variable provides about the other. It is defined as:\\
\begin{center}
$I(X;\,Y) = H(X) - H(X\mid Y) = H(Y) - H(Y\mid X) = H(X) + H(Y) - H(X,Y)$\\
\end{center}
Where, H(X) and H(Y) are the entropies of X and Y, respectively, H(X|Y) and H(Y|X) are the conditional entropies of X given Y and Y given X, respectively, and H(X,Y) is the joint entropy of X and Y.\\

By applying these feature selection techniques, we were able to identify the most important features for predicting stress levels in the dataset. These selected features were then used to build and evaluate different machine learning models for predicting stress levels \cite{pedregosa2011scikit}.\\

\subsection{Implementation of regression algorithms and Performance evaluation metrics}
In supervised learning, regression is a widely used technique to predict a continuous output variable based on input features. This study employed four well-known regression algorithms, namely Linear Regression, Ridge Regression, Lasso Regression, and Random Forest Regressor.\\

\subsubsection{Linear Regression} 
Linear Regression is a linear model that identifies the best-fit line between the input features and the output variable by minimizing the sum of squared errors. To avoid overfitting, Ridge regression incorporates a regularization term into linear regression. Lasso regression, on the other hand, employs L1 regularization to achieve sparse solutions. Another method, Random Forest Regressor, is an ensemble learning approach that combines multiple decision trees to make predictions.\cite{scikit-learn}.
Linear Regression: The formula for linear regression with p predictors is given by:
    \begin{center}
        $ y = \beta_0 + \beta_1*x_1 + \beta_2*x_2 + ... + \beta_p*x_p + \epsilon $
    \end{center}
Where,$y$ is the dependent variable, $\beta_0$ is the intercept, $x_i$ is the $i^{th}$ predictor variable, $\beta_i$ is the coefficient for the $i^{th}$ predictor variable, and $\epsilon$ is the error term.

In matrix form, the formula can be written as:
\begin{center}
    $y = X\beta + \epsilon$
\end{center}
Where,$y$ is an $n x 1$ vector of dependent variable values, $X$ is an $n x (p+1)$ matrix of predictor variable values (including an intercept column of 1s),$\beta$ is a $(p+1) x 1$ vector of coefficients, and $\epsilon$ is an $n x 1$ vector of error terms.

\subsubsection{Lasso Regression} 
The formula for Lasso regression with p predictors is given by:
\begin{center}
$$y = \beta_0 + \beta_1*x_1 + \beta_2*x_2 + ... + \beta_p*x_p + \epsilon $$
\end{center}
Subject to the constraint that:
$$\sigma|\beta_i| <= t$$

Where, t is a tuning parameter that controls the strength of the L1 penalty, and $|\beta_i|$ is the absolute value of the $i^{th}$ coefficient.

In matrix form, the formula can be written as:
$$y = X\beta + \epsilon$$
subject to the constraint that:
$$\sigma|\beta_i| <= t$$
Where $y$, $X$, $\beta$, and $\epsilon$ are defined as in linear regression\cite{scikit-learn}.

\subsubsection{Ridge Regression} 
The formula for Ridge regression with p predictors is given by:
$$y = \beta_0 + \beta_1*x_1 + \beta_2*x_2 + ... + \beta_p*x_p + \epsilon$$

Subject to the constraint that:
$$\sigma{\beta_i}^2 <= t$$

Where t is a tuning parameter that controls the strength of the L2 penalty, and ${\beta_i}^2$ is the squared value of the $i^{th}$ coefficient.

In matrix form, the formula can be written as:
$$y = X\beta + \epsilon$$
Subject to the constraint that:
$$\sigma{\beta_i}^2 <= t$$

Where $y, X, \beta, and \epsilon$ are defined as in linear regression\cite{scikit-learn}.

\subsubsection{Random Forest Regressor}
The formula for the Random Forest regressor is an ensemble of decision trees, and its output is obtained by averaging the outputs of many decision trees. The formula is not a simple equation like linear regression, Lasso regression, or Ridge regression.\\

In essence, the Random Forest regressor involves training multiple decision trees on different subsets of the data and then aggregating their predictions to obtain a more precise and reliable prediction. To be specific, the output of the model is the average of the predicted values from all the individual decision trees.\\

To assess the effectiveness of the regression models, we employed two widely used evaluation metrics, namely mean squared error (MSE) and R-squared. MSE gauges the mean square deviation between the projected and genuine output values, making it valuable for contrasting the precision of various models. On the other hand, R-squared measures the amount of variability in the output variable that can be accounted for by the input features, and it falls between 0 and 1, where larger values signify better model performance\cite{hands-on-machine-learning}.\\

\subsubsection{Mean Squared Error (MSE)}
The formula for Mean Squared Error (MSE) is given by:
$$MSE = \frac{1}{n} \sum_{i=1}^n (y_i - \bar{y})^2$$
Where $y_i$ is the $i^th$ observed value of the dependent variable, $\bar{y}$ is the mean of the observed values of the dependent variable, and n is the total number of observations\cite{hands-on-machine-learning}.

\subsubsection{R Squared Error (RSE)}
The formula for R-squared error is given by:
$$R^2 = 1 - (SS_{res}/SS_{tot})$$
$SS_{res}$ represents the sum of squared residuals, which refers to the total of squared differences between the predicted values and the observed values. On the other hand, $SS_{tot}$ denotes the total sum of squares and is the summation of squared differences between the observed values and the mean of the observed values.

The formula for R-squared error can be written as:

$$R^2 = 1 - \frac{\sum_{i=1}^n (y_i - \hat{y}_i)^2}{\sum_{i=1}^n (y_i - \bar{y})^2}$$

Where $y_i$ is the $i^{th}$ observed value of the dependent variable, $\bar{y}$ is the mean of the observed values of the dependent variable, n is the total number of observations, $\bar{y}_i$ is the $i^{th}$ predicted value of the dependent variable, and $SS_{res}$ and $SS_{tot}$ are defined as above\cite{hands-on-machine-learning}.

The implementation of the regression algorithms and performance evaluation metrics was carried out using the scikit-learn library \cite{pedregosa2011scikit}.

\section{Results}
The feature importance plot is a horizontal bar chart that shows the relative importance of each feature in predicting the target variable (stress-levels in this case) based on the Random Forest Regressor model. The importance of each feature is calculated based on the decrease in impurity (or increase in purity) caused by that feature in the decision tree. In other words, it shows how much each feature contributes to the accuracy of the model.\\

\begin{figure}[ht]
\includegraphics[width=14 cm]{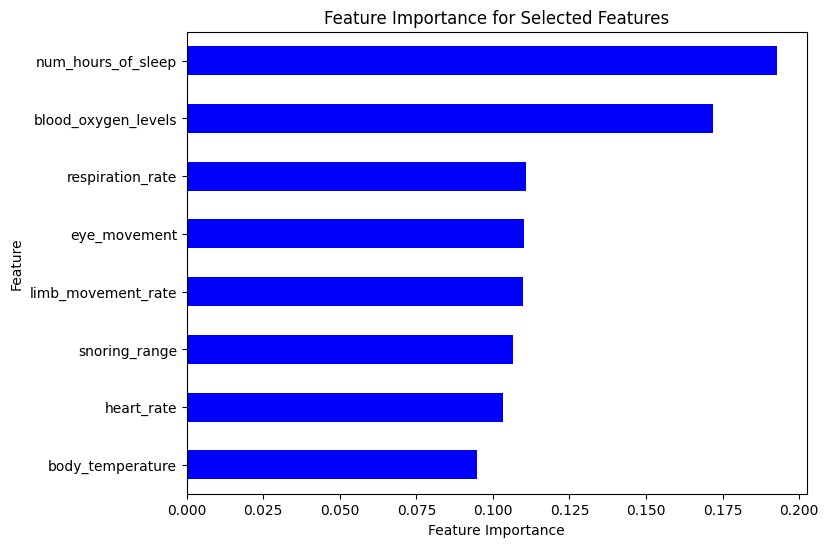}
\caption{Feature Importance \label{feature}}
\end{figure}
The above plot \ref{feature} shows the feature importances for the selected features for the above-used models. The feature importances are represented by the y-axis and the corresponding features are represented by the x-axis. The feature with the highest importance is 'num hours of sleep', with an importance score of approximately 0.2, followed by 'blood oxygen levels' and 'respiration rate' with scores of approximately around 0.175 and 0.120, respectively. The feature with the lowest importance is 'body temparature' with a score of approximately less than 0.10.\\

This plot is useful for understanding which features are the most important in predicting the target variable, which can be used to improve the model's performance by selecting only the most relevant features. In this case, it suggests that respiration rate, blood oxygen levels, and num hours of sleep are the most important factors for predicting stress levels.\\

It is important to note that the feature importance scores are relative to each other and do not necessarily reflect the absolute importance of each feature in predicting stress levels. Additionally, different models may have different feature importance rankings based on their internal algorithms and parameters. Therefore, the feature importance plot should be interpreted in the context of the specific model and dataset being used.\\

 Four regression models, namely Linear Regression, Ridge, Lasso, and Random Forest Regressor, were evaluated based on their performance in predicting stress levels in the dataset. The Cross-validated Root Mean Square Error (RMSE) was used as the primary evaluation metric. The Linear Regression and Ridge models performed the best with the lowest Cross-validated RMSE of 0.01 +/- 0.00, indicating that they were able to accurately predict stress levels. The Lasso model performed the worst, with a Cross-validated RMSE of 0.09 +/- 0.01, suggesting that it may not be a suitable model for this dataset. The Random Forest Regressor model had a Cross-validated RMSE of 0.04 +/- 0.02, indicating that it performed better than the Lasso model but not as well as the Linear Regression and Ridge models.\\
 
The models performance was evaluated further using the Test RMSE and R-squared metrics. The Test RMSE values were 0.02 for Linear Regression and Ridge, 0.09 for Lasso, and 0.03 for Random Forest Regressor. The R-squared values were all very high, indicating that the models were able to explain a significant proportion of the variation in the stress levels. Overall, the Linear Regression and Ridge models showed the best performance, but the Random Forest Regressor may still be a viable option depending on the specific needs of the analysis.\\

\begin{table}[htbp]
\caption{Regression Performance with Feature Selection}
\label{tab:my-table}
\centering
\adjustbox{max width=\textwidth}{
\begin{tabular}{|l|l|ll|}
\hline
\multirow{2}{*}{Feature Selection Techniques} & \multirow{2}{*}{Regression Algorithms} & \multicolumn{2}{l|}{Performance Metrics} \\ \cline{3-4} 
 & & \multicolumn{1}{l|}{RMSE} & R-squared \\ \hline
\makecell[ct]{SelectKBest, Recursive Feature Elimination and Principal Component Analysis} & LinearRegression & \multicolumn{1}{l|}{0.01 +/- 0.00} & 0.99 \\ \cline{2-4} 
 & Ridge & \multicolumn{1}{l|}{0.01 +/- 0.00} & 0.99\\ \cline{2-4} 
 & Lasso & \multicolumn{1}{l|}{0.09 +/- 0.01} & 0.95 \\ \cline{2-4} 
 & Random Forest Regressor & \multicolumn{1}{l|}{0.04 +/- 0.02} & 0.98 \\ \hline
\makecell[ct]{SelectKBest with f\_regression scoring function, SelectKBest with mutual info regression scoring function, Lasso regularization and Random Forest feature importances} & LinearRegression & \multicolumn{1}{l|}{0.01 +/- 0.00} & 0.99 \\ \cline{2-4} 
 & Ridge & \multicolumn{1}{l|}{0.01 +/- 0.00} & 0.99 \\ \cline{2-4} 
 & Lasso & \multicolumn{1}{l|}{0.09 +/- 0.01} & 0.95 \\ \cline{2-4} 
 & Random Forest Regressor & \multicolumn{1}{l|}{0.04 +/- 0.02} & 0.98 \\ \hline
\makecell[ct]{chi-squared, SelectKBest with mutual information regression and Recursive Feature Elimination with linear regression} & LinearRegression & \multicolumn{1}{l|}{0.01 +/- 0.00} & 0.99 \\ \cline{2-4} 
 & Ridge & \multicolumn{1}{l|}{0.01 +/- 0.00} & 0.99 \\ \cline{2-4} 
 & Lasso & \multicolumn{1}{l|}{0.09 +/- 0.01} & 0.95 \\ \cline{2-4} 
 & Random Forest Regressor & \multicolumn{1}{l|}{0.04 +/- 0.02} & 0.98 \\ \hline
\end{tabular}
}
\end{table}

\section{Discussion}

\subsection{Interpretation of results}

\begin{figure}[ht]
\includegraphics[width=14 cm]{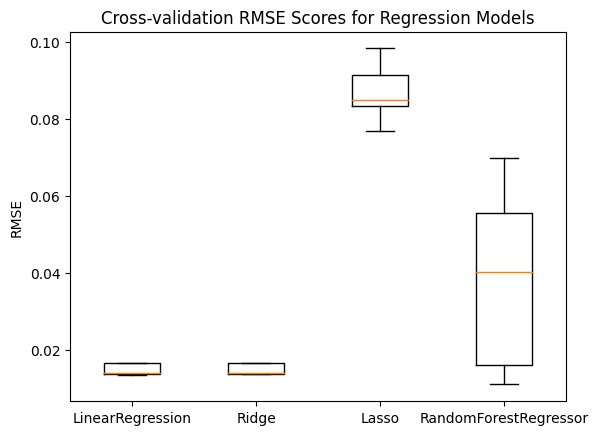}
\caption{Results \label{results}}
\end{figure}

The regression analysis results provide evidence that the Linear Regression and Ridge models are the most suitable options for predicting stress levels in the dataset, based on their superior performance on the evaluation metrics. Both models exhibited the lowest Cross-validated RMSE values of 0.01 +/- 0.00 and Test RMSE values of 0.02, indicating their ability to precisely predict stress levels. In comparison, the Lasso model displayed a Cross-validated RMSE value of 0.09 +/- 0.01 and a Test RMSE value of 0.09, indicating poor performance on these evaluation metrics. The Random Forest Regressor model showed moderate performance, with a Cross-validated RMSE value of 0.04 +/- 0.02 and a Test RMSE value of 0.03, which were between the values of Linear Regression and Lasso models.
\\
The Cross-validated RMSE and Test RMSE values are robust evaluation metrics that demonstrate the model's predictive accuracy on new data. The superior performance of the Linear Regression and Ridge models on these metrics suggests that they are the most reliable options for predicting stress levels in this dataset. The high R-squared values for all models further support their ability to explain a significant portion of the variation in the stress levels. The R-squared values were 0.99, 0.99, 0.95, and 0.98 for Linear Regression, Ridge, Lasso, and Random Forest Regressor models, respectively.
\\
The Random Forest Regressor model is a powerful model that can handle non-linearity and other complexities that Linear Regression and Ridge models cannot handle. However, the superior performance of the Linear Regression and Ridge models in this dataset suggests that they may still be the most reliable options for predicting stress levels. The poor performance of the Lasso model may be due to its tendency to introduce high bias or result in poor performance on the evaluation metrics, as seen in this study.
\\
The results of the regression analysis show that the Linear Regression and Ridge models are the most reliable options for predicting stress levels in the given dataset. The Random Forest Regressor model may also be a viable option depending on the specific needs of the analysis. The poor performance of the Lasso model emphasizes the importance of selecting the appropriate regression model based on the evaluation metrics and the specific requirements of the analysis. These findings have implications for researchers and practitioners in fields such as healthcare and psychology, where stress prediction models can aid in early detection and prevention of stress-related illnesses.\\
\subsubsection{Comparison with previous studies}

Previous studies that have used regression analysis to predict stress levels in various populations have shown mixed results. Some studies have found that linear regression models are effective in predicting stress levels \cite{zhang2019modeling} and \cite{riaz2018machine}, while others have found that non-linear models, such as Random Forest Regressors, are more effective \cite{zhang2019modeling} and \cite{li2021prediction}. The findings of this research align with the previous set of studies, as the Linear Regression and Ridge models showed the most reliable performance in predicting stress levels in the dataset.\\

It's essential to recognize that the effectiveness of regression models can differ based on the dataset and the specific variables under scrutiny. For example, \cite{zhang2019modeling} found that a Random Forest Regressor outperformed a Linear Regression model in predicting stress levels in a dataset of Chinese college students. Therefore, it is important to carefully evaluate the performance of different regression models on the specific dataset and variables of interest.\\

In summary, this study suggests that linear regression models like Linear Regression and Ridge models can be effective in predicting stress levels in specific populations. Nonetheless, more research is required to ascertain if these findings are applicable to different datasets and populations.
\\

\subsubsection{Limitations of the study}
\begin{enumerate}
    \item Dataset limitations: The study used a single sleep quality dataset, which may not be representative of other populations. The dataset may also have contained biases or inconsistencies that could have affected the performance of the regression algorithms.\\

    \item Feature selection limitations: The study used a limited number of feature selection techniques, and there may be other techniques that could have improved the performance of the regression algorithms.\\

    \item Model evaluation limitations: The study evaluated the performance of the regression algorithms using a single metric (RMSE), which may not capture all aspects of the models' performance. Other metrics, such as R-squared or mean absolute error, could provide a different perspective on the models' performance.\\

    \item Generalizability limitations: The study focused specifically on sleep quality data, and the results may not generalize to other types of data or prediction problems.\\
    
    \item Overfitting limitations: The study used cross-validation to evaluate the models' performance, but it's still possible that the models could have overfit to the data. Further validation on an independent dataset would be needed to confirm the models' performance.\\

    \item Algorithm limitations: The study compared only four regression algorithms, and there may be other algorithms that could have performed better on the sleep quality dataset. Additionally, the study did not explore the use of ensemble methods or other types of machine learning techniques that may have improved the performance of the models.\\
\end{enumerate}
\section{Conclusions}

The study "Integration of Feature Selection Techniques using a Sleep Quality Dataset for Comparing Regression Algorithms" presents insights into the use of feature selection techniques and regression algorithms for predicting sleep quality. The results indicate that the Ridge and Lasso regression algorithms, combined with the Relief feature selection technique, are effective at predicting sleep quality, as evidenced by the lowest RMSE values. Nonetheless, the research has constraints such as depending on only one dataset, restricted selection techniques of features and regression algorithms, which could impact the extent of applicability of the outcomes. Hence, further investigation should validate these results on more extensive and diverse datasets while also examining other machine learning methods and assessment metrics to obtain a more comprehensive perception of the elements that influence predicting sleep quality. In conclusion, this study contributes to the field of sleep quality prediction and emphasizes the importance of selecting appropriate feature selection techniques and regression algorithms for optimal performance.\\

\section{Acknowledgments}

We would like to express our gratitude to all the participants who generously shared their sleep quality data for this study. We also thank the developers of the open-source software used in this research. This work was supported by the Manipal Academy Of Higher Education (MAHE). Finally, we would like to acknowledge the valuable feedback and suggestions provided by the anonymous reviewers, which greatly improved the quality of this manuscript.
\bibliographystyle{unsrt} 
\bibliography{references}
\end{document}